\documentclass[10pt,twocolumn,letterpaper]{article}

\usepackage[pagenumbers]{cvpr} % To force page numbers, e.g. for an arXiv version

\usepackage{graphicx}
\usepackage{amsmath}
\usepackage{amssymb}
\usepackage{booktabs}
\usepackage[normalem]{ulem}
\usepackage{algpseudocode}
\usepackage{graphicx}
\usepackage{floatrow}
\usepackage{siunitx}
\usepackage{xcolor}
\usepackage{amsfonts}
\usepackage{eucal}

\usepackage[pagebackref,breaklinks,colorlinks]{hyperref}

\usepackage[capitalize]{cleveref}
\crefname{section}{Sec.}{Secs.}
\Crefname{section}{Section}{Sections}
\Crefname{table}{Table}{Tables}
\crefname{table}{Tab.}{Tabs.}

\def\cvprPaperID{12247} % *** Enter the CVPR Paper ID here
\def\confName{CVPR}
\def\confYear{2023}

\begin{document}

\title{ProtoVAE: Prototypical Networks for Unsupervised Disentanglement}

%\author{Vaishnavi Patil \small{*} \\
%University of Maryland\\
%College Park, MD\\
%{\tt\small vspatil@umd.edu}
%\and
%Matthew Evanusa \thanks{These authors contributed equally.}\\
%University of Maryland\\
%College Park, MD\\
%{\tt\small mevanusa@umd.edu}
%\and
%Joseph JaJa \\
%University of Maryland\\
%College Park, MD\\
%{\tt\small josephj@umd.edu}
%}

\author{\hspace{-4mm} Vaishnavi Patil$^{*}$ \hspace{-2mm}  \qquad Matthew Evanusa\thanks{These authors contributed equally.} \qquad  \hspace{3mm} Joseph JaJa\\
University of Maryland\\
College Park, MD\\
{\tt\small \{vspatil, mevanusa, josephj\}@umd.edu}
}
\maketitle

%%%%%%%%% ABSTRACT
\begin{abstract}
Generative modeling and self-supervised learning have in recent years made great strides towards learning from data in a completely \emph{unsupervised} way. There is still however an open area of investigation into guiding a neural network to encode the data into representations that are interpretable or explainable. The problem of unsupervised \textit{disentanglement} is of particular importance as it proposes to discover the different latent factors of variation or semantic concepts from the data alone, without labeled examples, and encode them into structurally disjoint latent representations. Without additional constraints or inductive biases placed in the network, a generative model may learn the data distribution and encode the factors, but not necessarily in a disentangled way. Here, we introduce a novel deep generative VAE-based model, ProtoVAE, that leverages a deep metric learning Prototypical network trained using self-supervision to impose these constraints. The prototypical network constrains the mapping of the representation space to data space to ensure that controlled changes in the representation space are mapped to changes in the factors of variations in the data space. Our model is completely unsupervised and requires no \textit{a priori} knowledge of the dataset, including the number of factors. We evaluate our proposed model on the benchmark dSprites, 3DShapes, and MPI3D disentanglement datasets, showing state of the art results against previous methods via qualitative traversals in the latent space, as well as quantitative disentanglement metrics. We further qualitatively demonstrate the effectiveness of our model on the real-world CelebA dataset. 

% to follow the established constraints of local isometry as well as to enforce changes in its latent representation remain distinct, consistent and noticeable. \mse{These constraints are imposed by changing the VAE representation space in a controlled manner via the process of \textit{interventions} in a self-supervised dataset generation process.}

% n both the encoding of the factors and the generating of data from these factors. 

%We introduce a novel deep generative VAE-based model, ProtoVAE, that leverages a deep metric learning Prototypical network trained using the generated self-supervised data 
%the incorporation of inductive biases in the problem definition that allow a neural network to learn to disentangle by generating its own data, in a self-supervised manner. 
\end{abstract}

%%%%%%%%% BODY TEXT

\section{Introduction}
\label{intro}

One theory of the success of deep learning models for supervised learning revolves around their ability to learn mappings from the input space to a lower dimensional abstract representation space which are best predictive of the corresponding labels \cite{tishby2015deep}. However, for the models to be robust to noise and adversarial examples, be transferable to different domains and distributions and interpretable, we need to impose additional constraints on the learning paradigm. As a promising solution to this, the models can be encouraged to focus on \textit{all} the latent "distinctive properties" of the data distribution and encode them into a representation for downstream supervised tasks. These latent distinctive properties or \textit{factors of variations} are the interpretable abstract concepts that describe the data.
% \mse{However, additional requirements of the network, such as being robust to noise and adversarial examples, being transferable to different domains, and interpretability, require us to impose certain constraints on the network architecture and learning.} \sout{However, additional constraints are necessary for these representations to be robust to noise and adversarial examples, transferable to different domain settings and distribution shifts and interpretable for explainable predictions.} \mse{One approach to guiding the network is to ecourage it to focus on the "distinctive properties" of the data distribution, and encode each property into unique representations, representations which are then used for downstream tasks.} These properties are known as \textit{factors of variation}, and are the interpretable abstract concepts present in the data. \sout{ As a promising solution to this, the networks can be encouraged to encode each latent distinctive property of the data distribution into a representation and then use these representations for downstream tasks. These latent distinctive properties or \textit{factors of variations} are the interpretable abstract concepts that are present in the data.} 
The intuitive notion of \textit{disentanglement}, first proposed in \cite{bengio2013deep}, proposes to discover all the different factors of variations from the data, and encode each factor in a separate subspace or dimension of the learned latent representation. These disentangled representations are not only interpretable and give valuable insights into the data distribution but are also more robust for multiple downstream tasks \cite{bengio2013deep, schoelkopf2012causal} which might depend only on a subset of factors \cite{suter2019robustly}.

The problem of learning these disentangled representations in a completely \textit{unsupervised} way is particularly challenging as we do not have access to the ground truth labels of factors nor are privy to the true number of factors or their nature. Recent works have proposed to solve this problem by training generative networks to effectively model the data distribution and in turn the factors of variations. From this generative perspective of disentanglement, higher dimensional data is assumed to be a non-linear mapping of these factors of variation, where each factor assumes different values to generate specific examples in the data distribution. \cite{locatello2019challenging} intuitively characterizes representations which encode the factors as \textit{disentangled} if a change in a single underlying factor of variation in the data produces a change in a single factor of the learned representation (or a change in the subspace of the representation that encodes that factor). Conversely, from the generative perspective, for a representation to be disentangled, a change in a single subspace of the learned representation, when mapped to the data space, must produce a change in a single factor of variation.
% \mse{However, this is likely the situation that we care about in application, as we will likely not have labels for each ground truth factor of variation, and would like a model that discovers these on its own.} 

For this generative mapping between changes in the representation space to the changes in the factors of variations (in the data space) to be injective, we propose constrains on the changes in the factors of variations for pre-determined changes in the representation space. Each separate subspace of the representation, when changed, must map to a change in a \textit{unique} factor of variation which in turn encourages information about the different factors to be encoded in separate subspaces of the representation. Moreover, each separate subspace must \textit{consistently} map to a change in a single factor throughout the subspace range. This encourages the different subspaces of the representation to encode information only about a single factor of variation. 
The recent work of \cite{horan2021unsupervised} also demonstrated empirically that the concept of \textit{local isometry} was a good inductive bias for unsupervised disentanglement, and it can aid generative models in discovering a ``natural" decomposition of data into factors of variation. This local isometry constraint on the mapping enforces the changes in the data space to be proportional to any changes made in the representation space. In order to effectively impose the above constraints in an unsupervised manner, we turn towards deep metric learning. 

In recent years, metric learning has emerged as a powerful unsupervised learning paradigm for deep neural networks, in conjunction with self-supervised data augmentation. One of the more successful metric learning models, Prototypical Networks, projects the data into a new metric space where examples from the same class cluster around a prototype representation of the class and away from the prototypes of other classes. We use this ability of the network to cluster the different changes in the data space mapped by the corresponding changes in the representation space and thereby enforce the above described constraints.
We develop a novel deep generative model, ProtoVAE, consisting of a Prototypical Network and Variational Autoencoder network (VAE). The VAE acts as the generative component, while the Prototypical Network guides the VAE in separating out the representation space by imposing the constraints for disentanglement.

To learn these representations in an unsupervised way, as the prototypical network needs labeled data for clustering, we train the prototypical network using generated self-supervised datasets. To produce the self supervised dataset, we perform \textit{interventions} in the representation space, which change individual elements of the latent space and map the intervened representations to the data space. Owing to the self-supervised training, our model is able to disentangle without any explicit prior knowledge of the data, \textit{including} the number of desired factors.% \sout{and forces the model to learn on its own to separate out the factors.}

% These transformations in the latent representation space are done via the approach of \textit{interventions} (Sec \ref{self_supervised_data_generation}). We call the changes in the data space induced by interventions in the latent space \textit{actions}. 
% \sout{The prototypical network's main role is to detect and isolate \vsp{changes in the different factors and encode them in distinct regions of the metric space} 
% unique changes in the discovered generative factors resulting from intervention. This process allows us to perform disentanglement without knowledge of the underlying data distribution.}

%\sout{and propose building a novel deep generative neural network model for unsupervised disentanglement that is trained end-to-end with these constraints in mind. If we want to learn these factors in an unsupervised manner the model must be able to generate \mse{factor-specific} {these kinds of} changes in the data, as well as be able to discern the changes. }

%\sout{In recent years, metric learning has emerged as a powerful unsupervised learning paradigm for deep neural networks, in conjunction with self-supervised data augmentation.  Here we take a similar inspiration, and combine aspects of a Prototypical Neural Network with a self-supervised data generation mechanism that identifies and isolates isolated changes in the factors of variations and the corresponding representations.} 

In this work, our core contributions are:
\begin{itemize}
    \item We design a self-supervised data generation mechanism using a VAE that creates new samples via a process of intervention to train a metric-learning prototypical network.
    \item We design and implement a novel model, ProtoVAE, which combines a VAE and prototypical network to perform disentanglement without any prior knowledge of the underlying data.
    % \item ProtoVAE combines the self-supervised data generation mechanism of a VAE and discriminator network with a prototypical network 
    \item We empirically evaluate ProtoVAE on standard benchmark DSprites, 3DShapes, MPI3D, and CelebA datasets, showing state of the art results.
\end{itemize}

\section{ProtoVAE}
Our proposed model consists of a VAE \cite{kingma2014autoencoding, rezende2014stochastic} as the base generative model (Section \ref{sec::vae}). The VAE consists of an inference network which encodes the data into lower dimensional latent representations and a generator network that maps the representations back into the data space. To implicitly encourage the inference network to encode disentangled representations, we impose constraints on the generative mapping from changes in the representation space to changes in the factors of variations in the data space. This generative mapping is determined by both the generator and the inference networks. To generate self-supervised data for the prototypical network, we perform interventions (Sec \ref{self_supervised_data_generation}) which changes individual dimensions of the representation. Given a batch of latent representations encoded by the inference network, we first intervene on a dimension of the representation by changing its value to the value of another representation from the batch for the same dimension. The original representations and the intervened representations are then mapped into the data space by the generator network and concatenated to form a pair of original data and generated data from interventions. Given that the original and the intervened representations differ in a single dimension, the generative mapping should be constrained to ensure that the corresponding pair of original and generated data differs only in a single factor of variation. 

This constraint is enforced using a Prototypical network (Appendix A.2) which based on the idea that there exist an embedding in which examples from the same class cluster around a prototype representation for that class. Our proposed prototypical network (Section \ref{sec::protonet}) takes as input pairs of data generated by the self-supervised process described above, and maps these pairs of data into a metric space in which pairs generated by intervening on the same dimension cluster together. These clusters which are identifiable with intervening dimensions in-turn become identifiable with the factors of variation that differ in value between the pair when a dimension is intervened upon.
We further augment the prototypical network with a separate output head that enforces local isometry, by predicting the difference in the value of the intervened dimensions from the pair in the data space. Fig \ref{fig:arch} gives the diagram overview of the complete model.

% \sout{These constraints in the data space are enforced using a Prototypical network which clusters pairs of data based on the differing factor of variation into distinct regions of the metric space}. \suggest{A Protoypical Network component (Sec \ref{sec::protonet}) is fed in reconstructions of data resulting in changes of these different dimensions of the representation space, the gradient of which is fed backward to the entire model as a constraint during training.} Network With each cluster identifiable with a distinct factor of variation, clusters are in-turn identifiable with the dimension of the intervened representation. \mse{Fig \ref{fig:arch} gives the diagram overview of the complete model.}

Lastly, for the intervened representations to be mapped into the data space such that only a \textit{factor of variation} is changed, we constrain the generated data to lie in the true data distribution. This constraint can be effectively enforced in the representation space by minimizing the distance between the distribution of the original representations of the inference network and the intervened representations such that the generator network maps both the distributions to the true data distribution. We do so by training a discriminator network (Section \ref{sec::disc}) in the representation space to distinguish between the original and the intervened representations. The inference network which generates the original representations is then trained to fool the discriminator thus effectively bridging the distance between the distributions.

\subsection{Variational Autoencoder}
\label{sec::vae}

The base generative model consists of an inference network $q_\phi: \mathbb{R}^D \rightarrow \mathbb{R}^{d}$ that encodes the data $x$ to a lower dimensional representation $z$ and a generator network $p_\theta: \mathbb{R}^d \rightarrow \mathbb{R}^D$ which reconstructs the data $\hat{x}$ from the representations $z$. The inference and the generator network are trained together to maximize the evidence lower bound (ELBO) of the data log-likelihood as in eq. \ref{eq:elbo}.
\begin{equation} 
    \max_{\theta, \phi}\mathcal{L}_{V}(\theta, \phi) = \mathbb{E}_{q_{\phi} (z|x)} [\log p_\theta (x|z)]
    - \text{KL} (q_{\phi} (z|x) || p (z))
\label{eq:elbo}    
\end{equation}

Maximizing the first term of eq. \ref{eq:elbo} ensures that the latent representation encodes all the information needed to faithfully reconstruct the data from the representation alone. This ensures that the representations encode all the different factors of variations in the data. The KL divergence term creates an information bottleneck which enforces optimal, compact encoding of the data by enforcing the posterior distribution to be similar to the independent, non-informative prior distribution. For more details please refer to Appendix A.1. 

\subsection{Self-Supervised Data Generation}
\label{self_supervised_data_generation}

\begin{figure}[t]
\vskip 0.2in
\begin{center}
\centerline{\includegraphics[width=\textwidth]{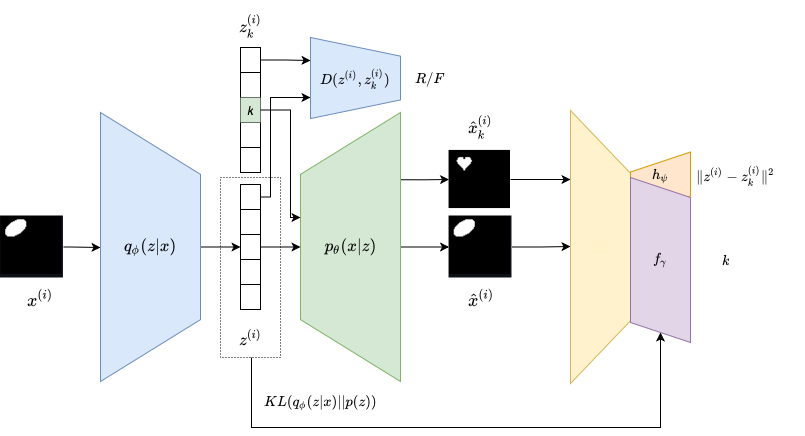}}
\caption{Architecture of our model consisting of a VAE, a discriminator and a Prototypical network. The representation $z$ from the inference network of the VAE (Sec \ref{sec::vae}), is changed at a particular dimension $k$ to get the intervened representation $\hat{z}_k$. A discriminator (Sec \ref{sec::disc}) is trained to distinguish between $z$ and $\hat{z}_k$ and the inference network is updated to fool the discriminator. $z$ and $\hat{z}_k$ are passed to the generator network to map it to the reconstructed data $\hat{x}$ and the intervened data $\hat{x}_k$. The original and the intervened data are concatenated to form the pair $(\hat{x},\hat{x}_k)$, which is then passed to the prototypical network. The prototypical network (Sec \ref{sec::protonet}) maps the pair closer to other pairs with the same dimension intervened. The prototypical network is updated by it's ability to correctly predict the intervened dimension of the query examples and the magnitude of the change $\|z-\hat{z}_k\|$.} 

\label{fig:arch}
\end{center}
%\vskip -2cm
\end{figure}

The prototypical network works to cluster changes in the factors of variations in the data space and provides gradients to the generator and inference network to better separate out the factors in the representation space. To do so, the prototypical network requires a set of supervised examples, called the support set, from each class to compute a prototype around which examples from the same class cluster. Furthermore, a supervised query set is required to compute the distance of query examples from the target prototypes and the subsequent loss is used to update the network. For learning disentangled representations in an \textit{unsupervised} way, we propose to generate these support and query sets using self-supervision. We describe the full algorithm in Appendix B.1.

Given a batch of data $x$, we first use the inference network to encode the data into the representation space $z \in \mathbb{R}^d$. Following \cite{suter2019robustly}, we define \textit{interventions} as the act of changing the value of a single dimension of the representation $k \in_R [d]$ while keeping the values of the other dimensions the same. We  change the value of the intervened dimension to another example's value for the same dimension. The result of intervening on representation $z$ in dimension $k$ produces the intervened representation $\hat{z}_k$. The representation and the intervened representation are then mapped by the generator network to $\hat{x}$ and $\hat{x}_k$ respectively. This pair of $(\hat{x}, \hat{x}_k)$ forms the input data for the prototypical network with the intervened dimension $k$ being the label and the difference in the representation space $|z - \hat{z}_k|$ as the label for the isometry head. The representation $z$ is intervened upon every dimension to generate $d$ support sets and is also intervened upon one dimension chosen uniformly from $[d]$ to generate the query set. The self-supervised data generation algorithm takes in a batch of data $x$ and outputs $d$ support sets $S=\{S_1, \cdots, S_d\}$, one query set $Q$, labels for the query set $L$ and labels for training the isometric head $I$.

\subsection{Prototypical Network}
\label{sec::protonet}

Our proposed prototypical network maps a pair of data generated by intervening on a single dimension of the representation to a lower dimensional metric space. By mapping \textit{a pair of data} to the metric space, the prototypical network can focus on the factor differing in value between the pair while being invariant to the values of the other non-differing factors. Critically, the factor differing in value remains the same across pairs of different examples when the same dimension is intervened upon and hence should be mapped closer in the metric space. Thus, comparing a pair of data allows the prototypical network to focus on the \textit{change} or \textit{difference} that was brought about by the intervened dimension, and makes the central focus of the losses this change.

The prototypical network first takes in elements of the generated support set $S$ (described in Section \ref{self_supervised_data_generation}) and computes an $m$-dimensional representation through the embedding function $f_\gamma : \mathbb{R}^D \times \mathbb{R}^D \rightarrow \mathbb{R}^m$. In this $m$-dimensional space, the prototypical network computes a prototype embedding $c_k$ for each element in the support set $S_k \in S$ using eq. \ref{eq:2}:
\begin{equation}  \label{eq:2}
    c_k = \frac{1}{|S_k|} \sum_{s_k^{(i)} \in S_k} f_\gamma (s_k^{(i)}) 
\end{equation}
While the support set is used to compute the prototypes, the query set is used to compute the loss by calculating the distance of its embeddings in the metric space to the target prototypes. For each dimension of the representation to encode information about a \textit{unique} factor of variation, each dimension when intervened upon and mapped to the data space must change a different factor of variation. Thus embeddings of pairs of data generated with the same intervening dimension of the representation must cluster closer in the metric space and away from the clusters of other dimensions. To enforce this, we introduce the \textit{uniqueness} loss which is computed for each query $q^{(i)}$ example by calculating the negative log-likelihood of the true class $l$ as in eq. \ref{eq:uniqueness}:
\begin{equation} \label{eq:uniqueness}
 \begin{aligned}
    &\min_{\gamma, \phi, \theta}\mathcal{L}_{U} (\gamma, \phi, \theta) = \\
    &- \frac{1}{|Q|} \sum_{q^{(i)} \in Q} \log p_\gamma (t=l | q^{(i)}) \cdot \text{KL} (q_{\phi} (z_l | x) || p(z))
    \end{aligned}
\end{equation}
where the probability of each class $p_\gamma(t=l | q^{(i)})$ is calculated as a distribution over the Euclidean distance $d$ to the prototypes as in eq. \ref{eq:dist}.
\begin{equation} \label{eq:dist}
p_\gamma (t=l | q^{(i)}) = \frac{\exp{(-d(f_\gamma (q^{(i)}), c_l))}}{\sum_{k'} \exp{(-d(f_\gamma (q^{(i)}), c_{k'}))}}
\end{equation}
The loss for every intervening dimension of the query examples is multiplied by the KL-divergence of that dimension, averaged for the batch of examples. This ensures that the loss for the intervening dimensions is scaled by amount of information encoded by that dimension. For the dimensions that do not encode any information, and hence do not change any factor of variation upon intervention, the corresponding loss is scaled by zero. This is important as we do not need any prior assumptions on the dimension of the representation needed to encode all the factors and the VAE can find the right number of dimensions needed.

In addition to the uniqueness loss, we want each dimension to consistently encode only a single factor of variation.  When the representation $z$ is first intervened on dimension $k$ and mapped to the data space it makes a certain change in factor.  When the representation is intervened again at dimension $k$ by a different amount and mapped to the data space it should produce a change in the same factor, irrespective of the amount it was changed. When passed in to the prototypical network, the pair of data generated by the original $z$ and intervened $\hat{z}_k$ must be embedded in the prototypical metric space closer to the pair generated by the representation intervened in the same dimension by a different amount. To enforce this we introduce the \textit{consistency} loss in eq. \ref{eq:consistency} where the prototypes are replaced by the embeddings of the same example in the support set. 
\begin{equation} \label{eq:consistency}
\begin{aligned}
    &\min_{\gamma, \phi, \theta}\mathcal{L}_{C} (\gamma, \phi, \theta) = \\ &- \frac{1}{|Q|} \sum_{q^{(i)} \in Q} \log r_\gamma (t=l | q^{(i)}) \cdot KL (q_{\phi} (z_l | x) || p(z))
\end{aligned}
\end{equation}
With $r_\gamma$ calculated as follows:
\begin{equation}
r_\gamma (t=l | q^{(i)}) = \frac{\exp{(-d(f_\gamma (q^{(i)}),f_\gamma(s_l^{(i)})))}}{\sum_{k'} \exp{(-d(f_\gamma (q^{(i)}), f_\gamma(s_{k'}^{(i)})))}}
\end{equation}
The consistency loss and the uniqueness loss are added together to get a combined prototypical loss eq. \ref{eq:proto_final}
\begin{equation} \label{eq:proto_final}
    L_{P} (\gamma, \phi, \theta) = L_{C} + L_{U} 
\end{equation}

As an additional inductive bias, as proposed in \cite{horan2021unsupervised}, we constrain the generative mapping between original and intervened representation $(z, \hat{z}_k)$ and the generated pair $(\hat{x}, \hat{x}_k)$ to be locally isometric \cite{Donoho5591}. Thus the factor changed in $\hat{x}_k$ when compared with $\hat{x}$ must differ in value proportional to the corresponding change in dimension $k$ of $z$ and the intervened $\hat{z}_k$. This serves as an imperative inductive bias for unsupervised disentanglement.

%\sout{Moreover, for a composition of interventions the corresponding generated data when compared with the original reconstructions must be isometric to each dimension of the representation.}

%\sout{Recent wok proposed by \cite{horan2021unsupervised} claim that local isometry \cite{Donoho5591} is one possible inductive bias that can help the model discover a "natural" decomposition into subgroups the actions of which correspond to the factors of variations of the data. Therefore we constrain the mapping of the interventions in the latent space to the intervention-observations to be isometric. Moreover, for a composition of intervention-transformations the corresponding intervention-observations when compared with the data must be isometric to each dimension of the representation.} 
The additional head $h_\psi: \mathbb{R}^D \times \mathbb{R}^D \rightarrow \mathbb{R}^d$, when given a pair of data, is trained to predict the difference in the values for the all the dimensions of $z$ and $\hat{z}_k$ through the loss function in eq. \ref{iso}.
\begin{equation} \label{iso}
    \min_{\psi, \theta, \phi} \mathcal{L}_{I} (\psi, \theta, \phi) = \|h_\psi((\hat{x}, \hat{x}_k)) - |z - \hat{z}_k| \|^2
\end{equation}
The training data for the isometry head generated in a self supervised manner as described in section \ref{self_supervised_data_generation} where the support set $S$ consists of data pairs and the set $I$ consists of the corresponding targets. In the final implementation, $f_\gamma$ and $h_\psi$ share all hidden convolutional layers and differ only in the final fully connected layer.

\begin{figure*}[ht!]
\centering
\begin{subfigure}{.47\textwidth}
  \centering
  \includegraphics[width=\textwidth]{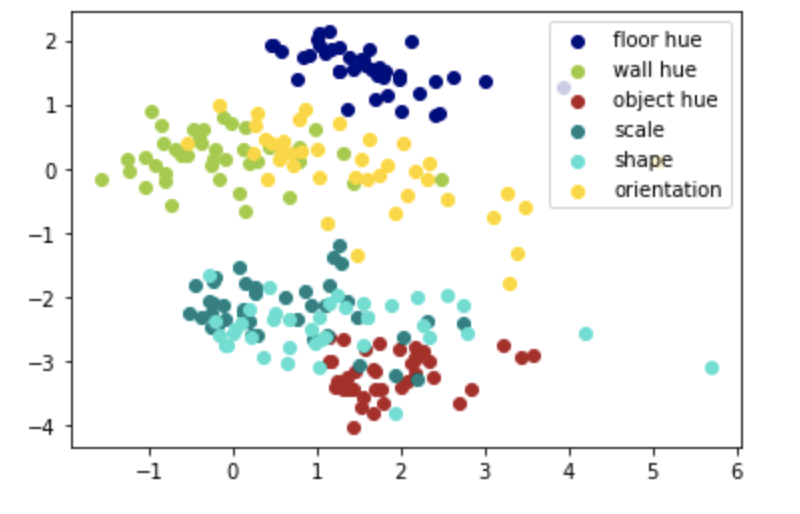}
  \label{fig:sub1}
\end{subfigure}%
\begin{subfigure}{.47\textwidth}
  \centering
  \includegraphics[width=\textwidth]{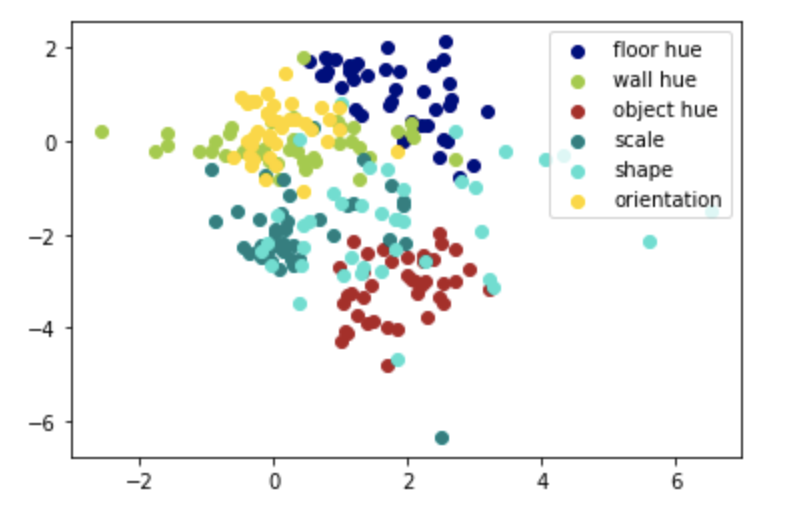}
  \label{fig:sub2}
\end{subfigure}
\caption{Output of the prototypical network with embedding dim $m=2$ when the input is real pairs of data from the 3DShapes dataset which differ in a single factor of variation. Each color corresponds to a unique factor which differs in value amongst the pair. The network clusters the changes correctly on the pairs from the original dataset. This suggests that the prototypical network is clustering pairs of images based on the changed factor of variation. \textit{Left}: $\lambda=10$. \textit{Right}: $\lambda=5$. }

\label{fig:prototypical_clustering}
\end{figure*}

\subsection{Representation Space Discriminator}
\label{sec::disc}

% \mse{Otherwise, the network may start to drift off-distribution as it generates training data for itself, which we observed in practice.}
%\sout{we need to enforce the distribution of the intervened latent representations to be close distribution of the original representations.} 

To restrict the realm of ``changes" in the data space, made by intervened representations, to only the factors of variation present in the dataset we regularize the intervened representations to map to data in the true data distribution. By minimizing the reconstruction loss in eq. \ref{eq:elbo}, the decoder learns to map the latent representations learned by the inference network $z \sim q_\phi (z)$ to the true data distribution $q(x)$. To enforce the decoder to also map the intervened representations $\hat{z} \sim q (\hat{z})$ to the true data distribution we propose to minimize the distance between distributions of the true representations $q_\phi (z)$ and the representations after intervention $q(\hat{z})$ by minimizing the KL divergence between the distributions $\text{KL} (q_\phi(z) || q (\hat{z}))$. To this effect we introduce a discriminator as proposed in \cite{kim2019disentangling} in the representation space which is trained to distinguishes samples from the two distributions by minimizing the loss in eq. \ref{eq:disc}).
% The above described process does not guarantee that the intervened representations will lie in the same distribution, as the original representations which can lead to the intervened representations being mapped in the data space away from the true data distribution. To minimize the distance between the distribution of the original representations $z$ and the intervened representations $\hat{z}$, we train a separate discriminator network in the representation space to distinguish the intervened representations from the original representations by minimizing the loss in eq. \ref{eq:disc}.
%\sout{to encourage the VAE Inference network to produce encodings that look indistinguishable from the intervened representations. This discriminator is similar to the one proposed in \cite{kim2019disentangling} to minimize the total correlation term KL$(q(z)||\prod_i q(z_i))$ The discriminator is trained to minimize the loss:}
\begin{equation} \label{eq:disc}
    \min_w \mathcal{L}_{D} (w) = - [\mathbb{E}_{\hat{z}}[\log(D_w (\hat{z}))] + \mathbb{E}_{z} [\log(1-D_w (z))]]
\end{equation}

% Note here that the real samples for the discriminator are the intervened representations and the fake samples are the original representations. Similar to the training of the generator in a Generative Adversarial Network (GAN) \cite{goodfellow2014generative},
The inference network, which generates the representations $z \sim q_\phi(z)$, is regularized to fool this discriminator by minimizing the loss in eq. \ref{eq:Inference network}. This encourages the inference network to encode data into representations whose individual dimensions can be intervened on and mapped to the same data distribution. 
\begin{equation}\label{eq:Inference network}
    \min_{\phi} \mathcal{L}_{E}(\phi) = \mathbb{E}_{z}[\log(1 - D_w (z))]
\end{equation}

The final objective (eq. \ref{eq:final}) of our method is a weighted sum of the different losses, and is optimized by the network parameters of the VAE and the prototypical network corresponding to each loss.
\begin{equation} \label{eq:final}
\begin{aligned}
    \min_{\phi, \theta, \gamma, \psi} \mathcal{L} = - \mathcal{L}_{V} (\phi, \theta) + \alpha \mathcal{L}_{E}(\phi) &+ \lambda \mathcal{L}_{P} (\gamma, \phi, \theta) \\ &+ \kappa \mathcal{L}_{I} (\psi, \phi, \theta)
\end{aligned}
\end{equation}
% The weights of the discriminator $w$ are optimized to minimize the objective $\mathcal{L}_{D} (w)$.

\begin{figure*}[h!]
\centering
\begin{subfigure}{.45\textwidth}
  
  \includegraphics[width=\textwidth]{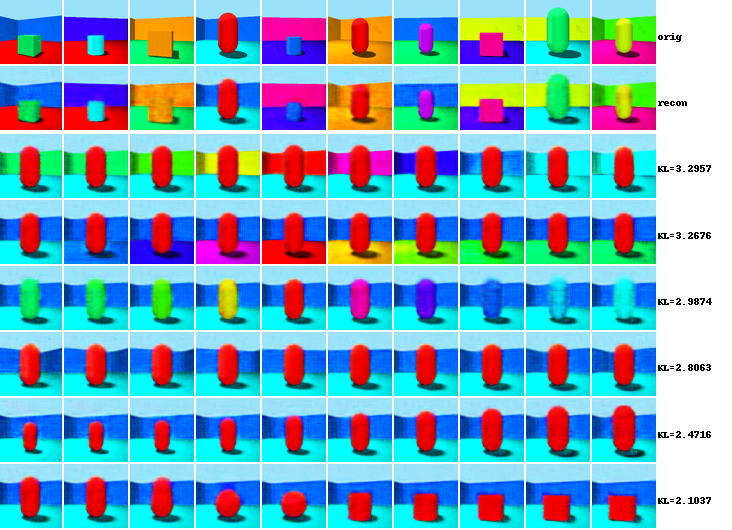}
  \label{fig:sub1}
\end{subfigure}%
\begin{subfigure}{.45\textwidth}
 
  \includegraphics[width=\textwidth]{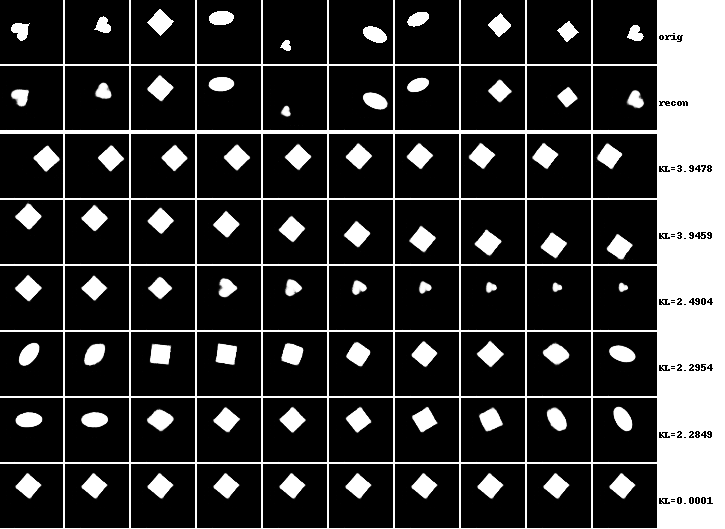}
  \label{fig:sub2}
\end{subfigure}
\caption{A comparison of latent traversals in latent space for the 3DShapes and Dsprites dataset. \textit{Left:} 3DShapes, \textit{Right: } Dsprites.  ProtoVAE produces smooth, disentangled latent representations. Row 1 and 2 are some sample original images, and their reconstructions generated by our model, respectively. Rows 3 downward are the traversals for each latent element, as detailed below. For 3DShapes, we actually see a near-perfect traversal across all of the known factors of variation.}

\label{fig:latent_trav}
\end{figure*}

\begin{figure}[h]
  \centering
  \includegraphics[width=.99\linewidth]{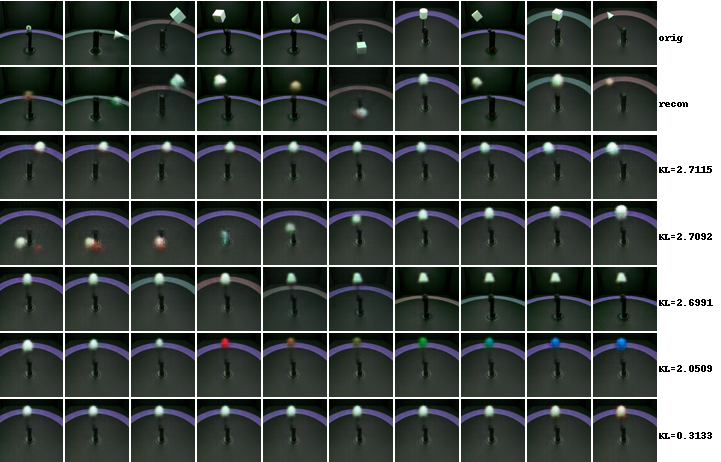}
  %\label{fig:sub1}
%\caption{}

\caption{Latent traversals on the MPI3D real world disentanglement dataset.  The data is collected via a camera that observes a jointed arm with known changed ground truth factors of variation.  From top to bottom: original data, reconstruction, arm angle left/right, arm angle top/bottom, background height, arm end color, size. The KL values represent the amount of information encoded by that dimension of the representation.}
\label{fig:mpi3d_traversal}
\end{figure}

\section{Empirical Evaluation}
To empirically evaluate our method, we perform both quantitative and qualitative evaluation on two synthetic datasets and one real dataset with known factors of variation and qualitative evaluation on CelebA dataset \cite{liu2015faceattributes}. The two synthetic and one real datasets are generated from independent ground truth factors of variation; DSprites \cite{DSprites17} binary 64 x 64 images with 5 factors of variation: 3 shapes, 6 scales, 40 orientations, 32 x-positions and 32 y-positions; 3D Shapes \cite{3DShapes18} 64 x 64 x 3 color images with 6 factors of variations: 4 shapes, 8 scales, 15 orientations, 10 floor colors, 10 wall colors and 10 object colors; MPI3D real \cite{NEURIPS2019_d97d404b} 64 x 64 x 3 color images with 7 factors of variations: 6 colors, 6 shapes, 2 sizes, 3 camera heights, 3 background colors, 40 horizontal axis, 40 vertical axis, and one real world dataset; CelebA. The details of the architecture for the different components and the corresponding hyperparameters are listed in the supplementary material (Appendix B.2) and (Appendix B.3) respectively.  
\begin{table*}[ht!]
\caption{Various disentanglement metrics evaluated across a number of state of the art methods for the DSprites and 3Dshapes dataset.  For all metrics, \textbf{higher is better.} The results for the other models are obtained using the hyperparameter settings and experimental conditions as described in \cite{locatello2019challenging}. The scores for all the models were averaged across ten runs with different random seeds, with standard deviation shown as $\pm$. Gr-FVAE is the GroupifyVAE variant applied to the FactorVAE, as this is the closest variant of the GroupifyVAE to our model and the results for which are taken from \cite{yang2021towards}. The highest values in a column are written in bold. As we see, the ProtoVAE outperforms the state of the art on a majority of the metrics. ProtoVAE hyperparameters for DSprites and 3Dshapes results shown are $\{\alpha=10, \lambda=10, \kappa=10\}$ and $\{\alpha=20, \lambda=20, \kappa=20\}$} 
\label{table_dsp}
% \vskip 0.15in
\begin{center}
\begin{small}
\begin{sc}
\begin{tabular}{lcccccccc}
\toprule
Datasets & & Dsprites & & & 3Dshapes & \\
\midrule 
Model & FVAE & DCI & MIG & FVAE & DCI & MIG\\%& Mod & Expl\\
\midrule
$\beta$-VAE   & 0.51 $\pm$ .10& 0.23 $\pm$ .10 & 0.15 $\pm$ .10 &  0.81 $\pm$ .10 & 0.44 $\pm$ .17 & 0.28 $\pm$ .18\\%& 0.76& 0.81\\
AnnVAE   &0.70 $\pm$ .10& 0.28 $\pm$ .10 & 0.23 $\pm$ .10 &  0.84 $\pm$ .09 &  0.46 $\pm$ .16 & 0.31 $\pm$ .15\\%& 0.76& 0.81\\
$\beta$-TCVAE & 0.68 $\pm$ .10 &  0.35 $\pm$ .06 & 0.17 $\pm$ .09 & 0.88 $\pm$ .07 &  0.63 $\pm$ .10 & 0.40 $\pm$ .18\\
FVAE & 0.74 $\pm$ .06 & 0.38 $\pm$ .10 & 0.28 $\pm$ .09 & 0.81 $\pm$ .06 & 0.47 $\pm$ .12 & 0.33 $\pm$ .14\\% 0.77 $\pm$ .01 & 0.81 \\% 0.77 $\pm$ .01 & 0.81  \\
Gr-FVAE & \textbf{0.75 $\pm$ .08} & 0.41 $\pm$ .07 & 0.31 $\pm$ .06 & 0.79 $\pm$ .06 & 0.49 $\pm$ .06 & 0.43 $\pm$ .11 \\
%\midrule
%IGAN$^*$     & 0.82 $\pm$ 0.01 & 0.60 $\pm$ 0.02 & 0.22 $\pm$ 0.01 & & 0.87 $\pm$ 0.01 &0.94 $\pm$ 0.01 & 0.82\\
%IGAN-CR$^*$      & 0.88 $\pm$ 0.01 & 0.71 $\pm$ 0.01 & 0.37 $\pm$ 0.01 & & 0.95 $\pm$ 0.01 &\textbf{0.96} & \textbf{0.85} \\
\midrule
%\textbf{ProtoVAE}   &    0.68 $\pm$ .07 & \textbf{0.46 $\pm$ .06} & \textbf{0.34 $\pm$ .09} &     \textbf{0.90 $\pm$ .06} & \textbf{0.84 $\pm$ .07} & \textbf{0.71 $\pm$ .11 } \\

\textbf{ProtoVAE}   &    0.70 $\pm$ .06 & \textbf{0.51 $\pm$ .04} & \textbf{0.37 $\pm$ .09} &     \textbf{0.90 $\pm$ .06} & \textbf{0.84 $\pm$ .07} & \textbf{0.71 $\pm$ .11 } \\%& 0.84 $\pm$ 0.02 &\textbf{ 0.85 }   \\
%Proto$^2$     &     0.88$\pm$ 0.02 & 0.74$\pm$ 0.02 & 0.40 $\pm$ 0.01& 0.47$\pm$ 0.01& 0.95$\pm$ 0.01 \\%& 0.82 $\pm$ 0.02& 0.84    \\
\bottomrule
\end{tabular}
\end{sc}
\end{small}
\end{center}
% \vskip -0.1in
\end{table*}

We qualitatively evaluate our model by intervening on the different dimensions of the learned representations and traversing the range of values of the dimension linearly in a fixed range $[-2,2]$. A model is better disentangled if the changes made in the data space while traversing a dimension are similar to the changes in a factor of variation in the data space.

% \begin{figure*}[ht]
% \begin{subfigure}{\textwidth}
%   \centering
%   \includegraphics[width=.5\linewidth]{WomanTraversal2.png}
%   \label{fig:sub1}
% \end{subfigure}%
% \begin{subfigure}{\textwidth}
%   \centering
%   \includegraphics[width=.5\linewidth]{WomanTraversal3.png}
%   \label{fig:sub2}
% \end{subfigure}
% \caption{A comparison of latent traversals in latent space for the CelebA celebrity photo dataset. Left: a latent dimension $z$ (maximum possible factors) of 16, Right: z equal to 10.  We show both to test if the model is agnostic to the maximum latent size, which our experiments appear to validate. Once the factors which contribute most to the reconstruction cost are discovered, the remaining latent dimensions collapse to the prior and are unused. Note for this case, there are a large number of unknown factors of variation, and the network is tasked with discovering factors in the real world.  Shown are traversals from two runs across various latent dimensions.  As we can see, ProtoVAE effectively discovers true, natural decompositions of the representation, many of which do correspond to factors of variation, many of which occur in both decompositions.  For example, hair color (row 5 left, row 4 right), skin tone (row 7, both), azimuth (row 8 left, row 3 right), and background color (row 3 left, row 1 right).}
% \label{fig:latent_trav_celeba}
% \end{figure*}

Our model both finds the correct number of factors and encodes them separately without any specific hyper-parameter tuning. From Figure \ref{fig:latent_trav} we can see that our method ProtoVAE, produces disentangled traversals covering both the number of factors as well as the entirety of the range of the values for the Dsprites and the 3DShapes dataset. The latent traversals on the MPI3D dataset can be found in figure \ref{fig:mpi3d_traversal}. Our method effectively separates the factors thus disentangling the learned latent representation without compromising on the reconstruction quality as seen from row 2. Owing to the unsupervised nature our method struggles to exactly disentangle the non-isometric discrete factor of shape in the DSprites dataset. For the 3DShapes dataset, in our traversals in figure \ref{fig:latent_trav}, we achieved near perfect disentanglement, completely unsupervised. In the Appendix D, we show the performance of the ProtoVAE for a subset of the Dsprites dataset with only a few factors. We show traversals from models FactorVAE and $\beta-$VAE, along with our model, with only a few isometric factors for comparison. Our proposed ProtoVAE is the only model that does not conflate two factors and encodes them in separate \mbox{dimensions} of the representation.

\begin{table}
\caption{Quantitative comparative metrics on the MPI3D dataset. ProtoVAE performs comparatively or better consistently across multiple metrics on a difficult real disentanglement dataset. See Fig. \ref{fig:mpi3d_traversal} for latent traversals on MPI3D. ProtoVAE hyperparameters for results shown are $\{\alpha=10, \lambda=2, \kappa=2\}$. The numbers for DisCo have been borrowed from their paper \cite{ren2021generative} for the VAE-based methods.}
 \begin{tabular}{lcccc}

\toprule
Model & FVAE & DCI & MIG \\%& Mod & Expl\\
\midrule
$\beta$-VAE  &  .41 $\pm$ .05 & .23 $\pm$ .04 & .06 $\pm$ .03  \\ %&  0.81 $\pm$ .10 & 0.44 $\pm$ .17 & 0.28 $\pm$ .18\\%& 0.76& 0.81\\
AnnVAE   & .29 $\pm$ .04 & .12 $\pm$ .02 & .07 $\pm$ .07 \\ %&  0.84 $\pm$ .09 &  0.46 $\pm$ .16 & 0.31 $\pm$ .15\\%& 0.76& 0.81\\
$\beta$-TCVAE & .45 $\pm$ .06 & .27 $\pm$ .03 & .16 $\pm$ .03  \\%& 0.88 $\pm$ .07 &  0.63 $\pm$ .10 & 0.40 $\pm$ .18\\
FVAE & .40 $\pm$ .04 & .30 $\pm$ .03 & \textbf{.23 $\pm$ .03} \\ %& 0.81 $\pm$ .06 & 0.47 $\pm$ .12 & 0.33 $\pm$ .14\\% 0.77 $\pm$ .01 & 0.81 \\% 0.77 $\pm$ .01 & 0.81  \\
DisCo & .39 $\pm$ .07 & .29 $\pm$ .02 & .07 $\pm$ .03 \\%& 0.79 $\pm$ .06 & 0.49 $\pm$ .06 & 0.43 $\pm$ .11 \\
%\midrule
%IGAN$^*$     & 0.82 $\pm$ 0.01 & 0.60 $\pm$ 0.02 & 0.22 $\pm$ 0.01 & & 0.87 $\pm$ 0.01 &0.94 $\pm$ 0.01 & 0.82\\
%IGAN-CR$^*$      & 0.88 $\pm$ 0.01 & 0.71 $\pm$ 0.01 & 0.37 $\pm$ 0.01 & & 0.95 $\pm$ 0.01 &\textbf{0.96} & \textbf{0.85} \\
\midrule
\textbf{ProtoVAE}  & \textbf{.46 $\pm$ .04} & \textbf{.38 $\pm$ .05} & \textbf{.25 $\pm$ .11} \\%&     \textbf{0.90 $\pm$ .06} & \textbf{0.84 $\pm$ .07} & \textbf{0.71 $\pm$ .11 } \\
\bottomrule

\end{tabular}
\end{table}

Furthermore, we quantitatively evaluate the learned representation by calculating state-of-the-art disentanglement metrics. We choose metrics from each of the three kinds of metrics described in \cite{zaidi2021measuring}; Intervention-based FactorVAE \cite{kim2019disentangling}, Predictor-based Disentanglement-Completeness-Informativeness (DCI) \cite{eastwood2018a} and Information-based Mutual Information Gap (MIG) \cite{chen2019isolating}. The metrics were implemented as proposed in \cite{locatello2019challenging} with the same hyperparameters. We refer the interested readers to \cite{zaidi2021measuring} for an intuitive understanding of the metrics. We highlight here that our model achieves a higher DCI metric and a higher corresponding \textit{completeness} and \textit{informativeness} metric, which reflects the mode covering capabilities of the learned \mbox{representation}. 

From table \ref{table_dsp} we see that on the DSprites dataset, our method outperforms the state of the art models in a majority of the metrics. Similar performance of our model on the 3DShapes dataset as seen in table. Also, the variance in the metrics for the different runs is significantly lower than of the previous methods, thus ensuring a more robust way to disentangle representations. For the real disentanglement dataset of MPI3D \cite{NEURIPS2019_d97d404b} consisting of a camera taking photos of an object attached to a jointed arm, we see that our model consistently either matches or outperforms the state of the art. From the baselines, especially important is the FactorVAE model, which is the base model upon which we add our contributions for the ProtoVAE model and hence use it as a comparison to demonstrate the effectiveness of our contributions.

% State of the art GAN methods, while being interpretable, suffer on this metric, and an initial hypothesis was that by adding a VAE, we gain the mode-covering capabilites of the VAE.  This hypothesis is verified by the high completeness scores across multiple trials.}

On the CelebA dataset, we find that across multiple runs, our model is able to find the same ``natural" decompositions that correspond to human-interpretable factors of variation consistently (Fig. 5). We notice that the model is not constrained to completely encode one factor per latent dimension and the model might encode different ranges of a factor in different latents; we see this occur for example when it encodes half of the azimuth in one latent, and half in another. However, as we can see, for the most part, each latent dimension contains information only about one factor of variation and even in the unsupervised regime our model still encodes natural decompositions.

We visualize the embedding space of a trained prototypical network using our method in Fig. \ref{fig:prototypical_clustering}. We see that when input to the prototypical network is pairs of images from the dataset, with one ground truth factor differing in value between the pair, the prototypical network effectively clusters the pairs based on the differing factor. This clustering aligns with the labels based on the intervened dimensions during training and thus points to the effectiveness of the prototypical network for encouraging disentanglement.

We also performed quantitative and qualitative ablation studies on the 3DShapes dataset by changing the values of $\alpha$, $\lambda$ and $\kappa$ to understand the effectiveness of each of the components and losses we introduce. The results of these ablations can be found in the supplementary material (Appendix C). Furthermore, we also perform ablation studies on the effect of dimension $m$ of the metric space of the prototypical network on the metric scores. We also show in the Appendix (Section C) some limitation cases where the representations of the model did not axis align with a few factors but was rotated with respect to those factors. We see that smaller values of the prototypical network metric space $m$ performs better by encoding data in tighter clusters which in turn it imposes stronger constraints on the VAE. The discriminator and the corresponding $\mathcal{L}_E$ helps in confining the encoding of the factors into a single dimension whereas $\mathcal{L}_P$ alone fails to do this effectively as seen in \ref{tab:}.  

\section{Related Works}
% Many state of the art approaches to disentanglement revolve around the use of VAEs or GANs to enforce the desired structure on the latent representations.  In the VAE approach, the network bottlenecks the input data into a lower dimensional space, forcing the network to (to the best of its ability) code the higher dimensional input in an efficient way. 
Many state-of-the-art unsupervised disentanglement methods extend the VAE objective function to impose additional constraints on the structure of the latent space to match the assumed independent factor distribution. 
$\beta$-VAE \cite{Higgins2017betaVAELB} and AnnealedVAE \cite{burgess2018understanding} heavily penalize the KL divergence term thus forcing the learned posterior distribution $q_{\phi}(z|x)$ to be independent like the prior. 
Factor-VAE \cite{kim2019disentangling} and $\beta$-TCVAE \cite{chen2019isolating} penalize the total correlation of the aggregated posterior $q_{\phi} (z)$. 
\begin{math}
TC = KL (q (z) || \prod_{i=1}^K q(z_i))
\end{math}
where the aggregated posterior is calculated as $q_{\phi} (z) = \mathbb{E}_{p (x)}[q(z|x_i)] = \frac{1}{N}\sum_{i=1}^N q_{\phi} (z|x_i)$ using adversarial and statistical techniques respectively. DIP-VAE \cite{kumar2018variational} forces the covariance matrix of the aggregated posterior $q(z)$ to be close to the identity matrix by method of moment matching. The changes that we described in the latent space are defined as intervention by \cite{suter2019robustly} to study the robustness of the learned representations under the Independent Mechanisms (IM) \cite{schoelkopf2012causal} assumption. Most closely related to our work are VAE models that learn to disentangle by altering the latent code.  In \cite{jha2018disentangling}, the authors use a VAE or AE with a split double latent code, with a cycle consistent loss, but required that attribute labels be known \textit{a priori}, which was also a requirement in \cite{feng2018dual} which learned by swapping out chunks of the latent code, and \cite{szabo2017challenges}, which used the labels as a constraint to find unique disentanglement. The authors in \cite{chen2020cyclically} also used cycle-consistent loss, but again required labeling.  In \cite{hu2018disentangling} the authors attempted unsupervised disentanglement with regular (non-variational) Autoencoder network models, by stacking one after another, our model instead uses a prototypical neural network.  In \cite{lezama2018overcoming} the authors derive a novel Jacobian loss combined with a student-teacher iterative training algorithm with an Autoencoder network model. In \cite{park2020swapping} the authors develop a latent-manipulating model aimed at human-interactive image manipulation tasks.

\begin{figure}[t!]
\vskip 0.2in
\begin{center}
\centerline{\includegraphics[width=.9\textwidth]{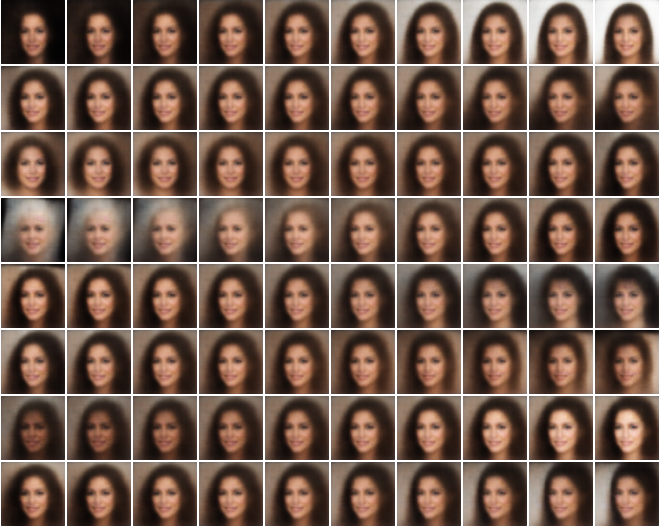}}
\caption{ Latent traversals on the CelebA dataset ProtoVAE successfully captures ground-truth factors of variation on real-world data. From top to bottom: background color, hairstyle, head angle, age, hairstyle, hair color, skin color, face profile.} 

\label{fig:arch}
\end{center}
%\vskip -2cm
\end{figure}

In \cite{yang2021towards} the authors use the group based definition by \cite{higgins2018definition} and a cycle consistency loss to define the elements of a group. Our work differs significantly as we do not re-encode the reconstructed data nor the generated data from interventions and instead use a prototypical network. \cite{zhu2021commutative} encode the the latent space of a VAE using the commutative Lie group and enforce constraints on the latent space. A recent work \cite{ren2021generative} propose to learn disentangled representations from pre-trained models using contrastive methods. 

The most prominent work from the GAN family is InfoGAN \cite{chen2016infogan} which learns disentangled, semantically meaningful representations by maximizing a lower bound on the intractable mutual information between the conditioning latent variables $c$ and the generated samples $G(z,c)$. InfoGAN-CR \cite{lin2020infogancr} and \cite{zhu2020learning} add a contrastive regularizer to the InfoGAN model to further encourage disentanglement. \cite{liu2020oogan} add orthogonal regularization to encourage independent representations.  However, all the GAN methods suffer from the limitation that they require \textit{a priori} the number of factors to be discovered, in addition to the number of values for all the discrete factors. For fairness of comparison, we thus only compare against methods that do not require these priors.
% , which is trained to predict the changes in the latent space given only the pairs of images generated from the respective latent codes.
%In practice, this is not ideal as discovering the number of factors, as well as having no \textit{a priori} knowledge of the structure of the factors, is a part of the disentanglement process. For this reason we do not compare these methods in our comparison table.}

\begin{table} \label{tab:alpha}
\caption{Ablation results for different values of $\alpha$, which shows that the discriminator helps in confining the encoding of the factors into a single dimension. This can be seen by a higher value of the $\beta$ VAE and the FactorVAE metrics for $\alpha=0$ but not for the MIG and the DCI metrics which require factors to be encoded in a single dimension.}
\begin{tabular}{lccc}
%\begin{center}
%\begin{sc}
%\begin{small}
\toprule
Metric & $ \alpha=0$ & $\alpha=10$ & $\alpha=20$ \\
\midrule
FVAE  & .93 $\pm$ .04 & .85  $\pm$ .03 & .88 $\pm$ .05 \\
DCI  & .78 $\pm$ .05 &  .81 $\pm$  .06 & .81 $\pm$ .07 \\
MIG  & .59 $\pm$ .07 &  .63 $\pm$ .04 & .65 $\pm$ .08 \\
$\beta$-VAE & .92 $\pm$ .06 &  .88 $\pm$ .04 & .90 $\pm$ .05 \\
\bottomrule
%\end{small}
%\end{sc}
%\end{center}
\end{tabular}
\end{table}

\section{Conclusion and Future Work}
In this work, we proposed a novel generative model consisting of a VAE and a Prototypical Network for learning disentangled representations in a completely unsupervised way, inspired by recent discovery of sufficient inductive biases. We impose constraints on the structure of the representations learned by training the model in an self-supervised manner to encode information about the different factors in separate dimensions of the representation. Our proposed method is able to outperform other state of the art networks on a number of metrics on three prominent disentanglement datasets. For future work, our method can be easily adapted to be trained in a weakly supervised regime with pairs of data differing in known number of factors being the prototypes for the prototypical network. The results can be possibly improved by intervening on multiple dimensions of the representations simultaneously. The importance of methods that can disentangle data \textit{without labels} is critical as data is plentiful and the resulting representations give interpretable insights into the variations in the data distribution, and can be used for downstream tasks. Our hope is this work adds evidence that self-supervised generative methods are important in this endeavor. 

%%%%%%%%% REFERENCES
{\small
\bibliographystyle{ieee_fullname}
\bibliography{egbib}
}

\end{document}